\documentclass[review]{elsarticle}

\usepackage[margin=2.5cm]{geometry}

\usepackage{lineno,hyperref}
\usepackage{wrapfig}
\usepackage{csquotes}
\usepackage{color}

\usepackage{array,multirow}
\usepackage{tabularx}
\usepackage{array}
\usepackage{booktabs}
\usepackage{multirow}
\usepackage{xcolor}
\usepackage{multirow}
\usepackage{array, makecell}
\usepackage{adjustbox}
\usepackage{braket}

\journal{ArXiv}

\begin{document}

\begin{frontmatter}

\title{Artificial Intelligence based Autonomous Molecular Design for Medical Therapeutic: A Perspective}

\author{Rajendra P. Joshi}

\author{Neeraj Kumar$^{*}$}
\cortext[mycorrespondingauthor]{Corresponding author}
\ead{neeraj.kumar@pnnl.gov}
\address{Pacific Northwest National Laboratory,
Richland, WA 99352, United States}

\begin{abstract}
 Domain-aware machine learning (ML) models have been increasingly adopted for accelerating small molecule therapeutic design in the recent years. These models have been enabled by significant advancement in state-of-the-art artificial intelligence (AI) and computing infrastructures. Several ML architectures are predominantly and independently used either for predicting the properties of small molecules, or for generating lead therapeutic candidates. Synergetically using these individual components along with robust representation and data generation techniques 
autonomously in closed loops holds enormous promise for  
accelerated drug design which is a time consuming and expensive task otherwise. In this perspective, we present the most recent breakthrough achieved by each of the components, and how such autonomous AI and ML workflow can be realized to radically accelerate the hit identification and lead optimization. Taken together, this could significantly shorten the timeline for end-to-end antiviral discovery and optimization times to weeks upon the arrival of a novel zoonotic transmission event. Our perspective serves as a guide for researchers to practice autonomous molecular design in therapeutic discovery. 

\end{abstract}

\end{frontmatter}

%\linenumbers

\section{Introduction}

Synthesizing and characterizing small molecules in a laboratory  with desired properties is a time consuming task\cite{DDR1_21_days}. %and tedious task with conventional methods. 
Until recently, experimental laboratories are mostly human operated; they relied completely on the experts of the field to design experiments, carry out characterization, analyze, validate, and conduct decision making for the final product. %  depending upon the quality of results obtained during the post experiment decision making. %  process. %Each of these steps is time consuming, expensive and resource intensive. 
Moreover,  the experimental process involves a series of steps, each requiring several correlated parameters that need to be tuned\cite{parameters_1, parameters_2},  
%-------    
%Such parameters can be unique depending upon the chemistry of materials, experimental setups, characterization techniques and their application field. Tuning these parameters requires careful exploration of parameter space, 
which is a daunting task, as each parameter set conventionally demands individual experiments. This has slowed down the discovery of high impact small molecules and/or materials, in some case by decades, with possible implications for diverse fields such as in energy storage, electronics, catalysis, drug discovery etc. Besides, the high impact materials of today come from exploring only a fraction of the known chemical space. Larger portions of the chemical space are still uncovered and it is expected to contain exotic materials with the potential to bring unprecedented advances to state-of-art technologies. Exploring such a large space with conventional experiments  will take time and lot of resources\cite{time, cost2, cost1, cost}. In this scenario, complete automation of laboratories is long overdue and have been used with limited success in the past\cite{loop3, loop4, loop2, loop}. 

%In most cases, the designs of experiments starts with screening the materials using the chemical intuitions and computational simulations such as density functional theory (DFT). DFT, although is the work horse of computational chemistry, it is  computationally challenging to use it in the huge known chemical space in the desired time.

Automating the computational design of molecules that integrates physics-based simulations and optimization with ML approaches are a feasible and efficient alternative instead;
%Computational approaches that use physics based simulations are efficient alternative instead 
it can contribute significantly in accelerating autonomous molecular design. 
%------------------------------------
%In this scenario, complete automation of laboratories is long overdue. The concept of laboratory automation is not new\cite{Lab-automation-1970}, but it was used with limited success for material discovery in the past.  Recently, it has re-emerged as the approach of interest due to the  significant development in computing architecture, sophisticated material synthesis and characterization techniques, and increasing successful adoption of deep learning based models in physical and biological science domains.
%------------------------------------
High throughput quantum mechanical calculations such as efficient density functional theory (DFT) based simulations are the first step towards this goal of providing insight into larger chemical space and have shown some promise %and are extensively and successfully used both computationally and experimentally
to accelerate novel molecule discovery. However, it still requires human intelligence for different decision-making processes and it cannot autonomously guide small molecule therapeutic discovery steps, thus slowing down the entire process. %are not autonomous. %with laboratory automation in mind. But such approach can not autonomously guide materials discovery, individually and require human intelligence for the decision making. 
%The concept of laboratory automation is not new, but it was used with limited success in the past. %This was limited because the experimental process. 
Additionally, inverse design of molecules is notoriously difficult  with DFT alone. %these approaches.
The amount of data produced by these high throughput methods      
is so large that it cannot be analyzed in real time with conventional methods.
Autonomous computational design and characterization of molecules %Such computational pipelines are 
is more important %[are only deemed approach]
in the scenarios where existing experimental/computational approaches are inefficient\cite{fragmentation, massSpectrometry}. 
One particular  example is the challenge associated with identifying new metabolites in a biological sample from mass spectrometry data, which requires mapping the fragmented spectra of novel molecules to the existing spectral library making it slow and tedious. In many cases, such references libraries do not exist, and a machine learning integrated, automated workflow could be an ideal choice to deploy for rapid identification of metabolites as well as to expand the existing libraries for future reference.
%Machine learning 
Such a workflow has shown the early ability to quickly screen molecules and accurately predict their properties for different applications. %This requires
The synergistic use of high throughput methods in a closed loop with machine learning based methods, capable of inverse design, %, to enable quick discovery of new lead molecules with tailored properties and accurate property predictions 
is considered vital for autonomous and accelerated discovery of molecules\cite{loop}. %Some researchers have successfully used this techniques in the laboratory and have started to gain some attention. 
In this perspective, we discuss how computational workflows for autonomous molecular design can guide the goal of laboratory automation and review the current state-of-the art artificial intelligence (AI) guided autonomous molecular design focusing mainly on small molecule therapeutic discovery.

\section{Components of Computational Autonomous Molecular Design Workflow}

%input data featurization is more critical than selecting a model. By a “descriptor” we mean any function that maps a molecule to a scalar value.

The workflow for computational autonomous molecular design (CAMD) must be an integrated and closed loop system with (i) efficient data generation and extraction tools, (ii) robust data representation techniques, (iii) physics based predictive machine learning models, and (iv) tools to generate new molecules using the knowledge learned from steps i-iii. Ideally, an autonomous computational workflow for molecule discovery would learn from its own experience and adjust its functionality as the chemical environment or the targeted functionality changes.  This can be achieved when all the components work  %Such systems can be realized by synergically using the proper representation of molecules along with efficient generative and predictive deep learning models working
in collaboration with each other, providing feedback while improving model performance as we move from one step to other.

\begin{figure}[h]
\begin{center}
\includegraphics[scale=0.4]{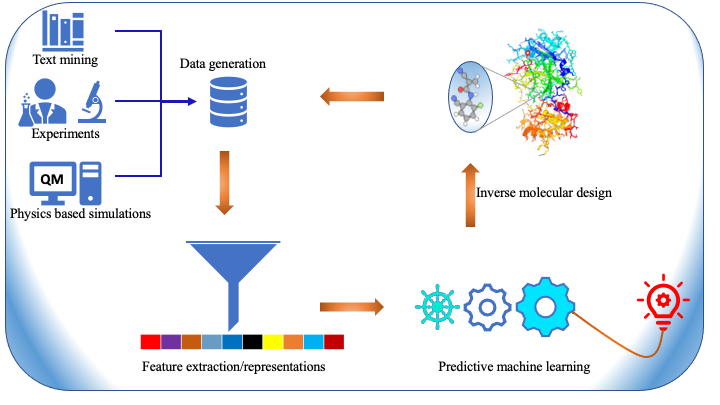}
\end{center}
\caption{Closed loop workflow for computational autonomous molecular design (CAMD) for medical therapeutics. Individual component of the workflow are labelled. It consists of data generation, feature extraction, predictive modeling and an inverse molecular design engine. }
\end{figure}

For data generation in CAMD, high-throughput density functional theory (DFT)\cite{DFT12, DFT23} is a common choice mainly because of its reasonable accuracy and efficiency\cite{jain2011high, DFT4}. In DFT, we feed in 3D-structures to predict the properties of interest. Data generated from DFT simulations is  processed to extract the more relevant data, which is then either used as input to learn the representation\cite{dft21, lee2020analytical}, or as a target required for the ML models\cite{QC_ML, chemical_accuracy23}. Data generated can be used in two different ways i.e.  to predict the properties of new molecules using a direct supervised ML approach, and to generate new molecules having desired properties of interest using inverse design. CAMD can be tied with supplementary components such as databases to  store the data and visualize it. The AI-assisted CAMD workflow presented here is the first step in developing automated workflows for molecular design. It will only accelerate the lead optimization. These workflows, in principle, can be combined with experimental setups for computer aided synthesis planning that includes synthesis and characterization tools at the cost of an increase in the complexity and expense. Instead, experimental measurements and characterization should be performed for only the lead compounds obtained from CAMD.

The data generated from inverse design in principle should be validated by using an integrated DFT method for the desired properties, or by docking with a target protein to find out its affinity in the closed loop system, then accordingly update the rest of the CAMD. These steps are then repeated in a closed loop, thus improving and optimizing the data representation, property prediction and new data generation component. Once we have confidence in our workflow to generate valid new molecules, the validation step with DFT can be bypassed or replaced with an ML predictive tool to make the workflow computationally more efficient. In the following, we briefly discuss the main component of the CAMD, while  reviewing the recent breakthroughs achieved.

%An ideal autonomous computational pipeline should be scalable and capable of real time processing of data without any human input and learn from its own experience and adjust its functionality as the chemical environment changes. 
%Such systems can be realized by synergically using the robot science with deep learning based models and high-through put approaches. 
%Such pipelines can be realized by integrating the separate modular components each of which performs a distinctive task. 

%synergically using the proper representation of molecules along with efficient generative and predictive deep learning models working in collaboration with each other, providing feedback while improving model performance as we move from one step to other. An autonomous molecular design pipeline should contain the efficient and effective representation scheme, property predicting and molecule generating tool. \textcolor{red}{FIGURE UNDER CONSTRUCTION !!!}

%First step for any experiment is the synthesis of materials. Every experiment has its own unique synthesis conditions which should be accurate as everything else depend upon this. These parameters in automated pipelines can be predicted by using machine learning based models, as has been successfully used before. This, however requires some data from previous experiments for training the machine learning model for accurate parameter prediction.

%\clearpage

\section{Data Generation and Molecular Representation} 

Machine learning models are data centric. Lack of accurate and well curated data %required for training the ML models 
is the main bottleneck limiting their use in many domains of physical and biological science. For some sub-domains, a limited amount of data exists that comes mainly from physics-based simulations in databases\cite{ramakrishnan2014quantum, qm99}, or from experimental databases such as NIST\cite{nist}. For other fields such as for bio-chemical reactions\cite{Modelseed}, we have databases with the free energy of reactions, but they are obtained with empirical methods, which are not considered ideal as ground truth for machine learning models. %Some unconventional yet very effective approaches have also been used to create data %in materials science  which otherwise are known to take lot of resources and time. 
For many domains, accurate and curated data does not exist. In these scenarios,
slightly unconventional yet very effective approaches of creating data from published scientific literature and patents for ML have recently gained adoption\cite{text_mined_data, textmining1, textmining2, textmining3}. %Published articles are rich source of information where valuable data comes in various forms, formats and places such as in text, tables, images, captions, supplementary documents.
%In a recent work, Ceder's et al. generated a large dataset of solid state synthesis reactions by text-mining data from published literature\cite{text_mined_data}, which can be used for machine learning synthesis reactions.
%Even for the field with some data, we have the issue of representation of molecules in machine learning model. 

%\section{Molecular representation in pipelines}
Robust representation of molecules is required for accurate functioning of the machine learning models\cite{representations1}. An ideal representation of molecules should be unique, invariant,  %with respect to different symmetry operations  
invertible, efficient to obtain, and should capture the physics, chemistry and structural motif of the molecules. 
Some of these can be achieved by using all the physical, chemical and structural properties\cite{MEGNet}, which all together are rarely well documented so getting this information is considered cumbersome task. % due to computational cost associated with extracting them. % for all the molecules known.
%Calculating such properties by doing physics based simulation is not always desired due to the computational cost.
Over time, this has been tackled by using several alternative approaches  that work well for specific problems\cite{representations2, augmentation, NMP, graphs1, graphs2, graphs3}. However, obtaining robust representations of molecules for diverse machine learning problems is still a challenging task and any gold standard method that works consistently for all kind of problems is yet to be  known. 
Molecular representations used in the literature falls into two broad groups, (a) 1D and/or 2D representations designed by experts using domain specific knowledge including properties from the simulation and experiments, and (b) iteratively learned molecular representations directly from the 3D nuclear coordinates/properties within ML frameworks.

\begin{figure}
    \centering
    \includegraphics[scale=0.35]{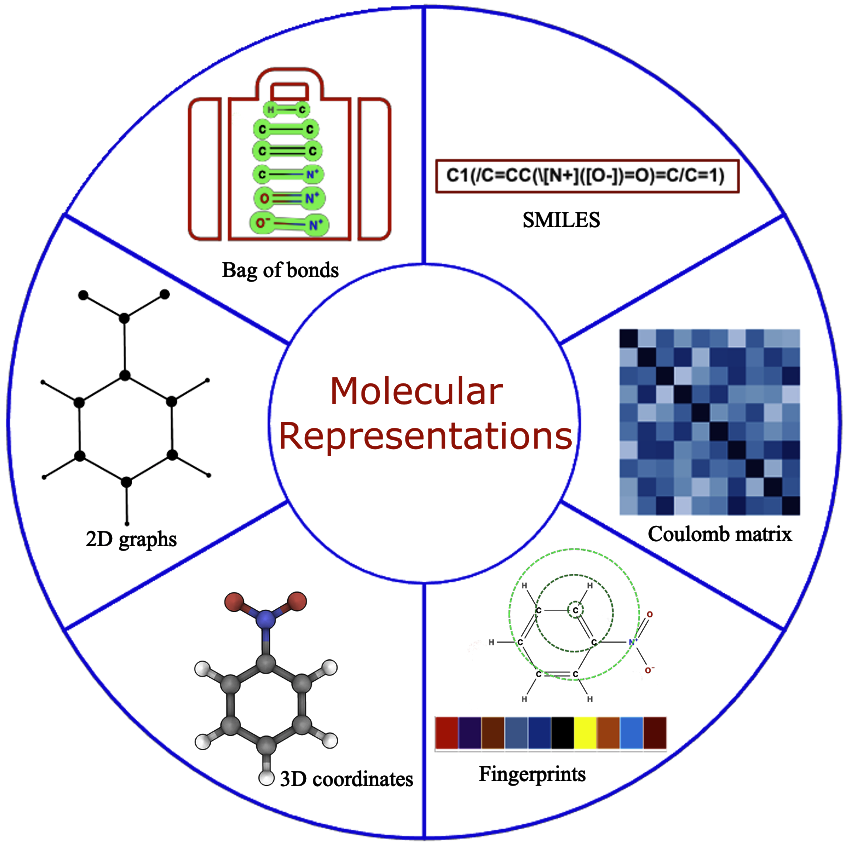}
    \caption{Different form of molecular representation used in the literature for predictive and generative modeling. }
    \label{fig:my_label}
\end{figure}

Expert engineered molecular representations have been extensively used for predictive modeling in the last decade which includes properties of the molecules\cite{coulomb1, bag-of-bonds}, structured text sequences\cite{smiles, inchi, inchi1} (SMILES, InChI),  molecular fingerprints\cite{fingerprint}  among others. Such representations are carefully
selected for each specific problem
using domain expertise, a lot of resources, and time. 
The SMILES representation of molecules is the main workhorse as a starting point for both representation learning as well as for generating expert engineered molecular descriptors. %Latter has been widely used by training the ML models in sequence of characters converting sequence 
For the latter, SMILES strings can be used directly as one hot encoded vector,  to calculate fingerprints or  to calculate the range of empirical properties using different open source platforms such as RDkit\cite{rdkit}, chemaxon\cite{chemaxon} thereby  by-passing expensive features generation from quantum chemistry/experiments  
by providing faster speed and  diverse properties, including 3D-coordinates, for molecular representations. 
Moreover, they can be easily converted into 2D-graphs, which is  preferred choice to date for generative modelling, where molecules are treated as graphs with  nodes and edges.
Although significant progress has been made in molecular generative  modeling using mainly SMILES strings\cite{smiles}, they 
often lead to generation of syntactically invalid molecules and are non-unique. In addition, they are also known to violate fundamental physics and chemistry based constraints\cite{selfies, blog}. Case specific solutions to circumvent some of these problems exist, but a universal solution is still unknown.
Extension of SMILES  were attempted by more robustly encoding  rings
and branches of molecules to find more concrete representations with high semantical and syntactical validity using canonical SMILES\cite{canoSMILES1, canoSMILES2}, InChI \cite{inchi, inchi1}, SMARTS\cite{smart},  DeepSMILES\cite{deepSMILES},  DESMILES\cite{desmiles} etc. More recently, Alan et al. proposed 100 \%  syntactically correct and robust string based representation of molecules known as SELFIES\cite{selfies},
which has been increasingly adopted for predictive and generative modelling\cite{selfiesApp1}.

Recently, molecular representations that can be iteratively learned directly from molecules have increasingly gained adoption, mainly for predictive molecular modeling achieving chemical accuracy for range of properties\cite{MEGNet, gschnet1, SchNetPack}. Such representations are more robust and out perform expert designed representations in drug discovery.\cite{dataset_size} %SMILES have been recently used as an input for learning the molecular representation using message passing neural networks\cite{NMP, mpn-app}. 
For representation learning, different variants of graph neural networks are a popular choice\cite{NMP, mpn-app}. It starts with generating atom (node) and bond (edge) features for all the atoms  and bonds within a molecule which are iteratively updated using graph traversal algorithms, taking into account the chemical environment information to learn a robust molecular representation. Starting atom and bond features of the molecule may just be one hot encoded vector to include only atom-type, bond-type or a list of properties of the atom and bonds derived from SMILES strings. Yang et al. achieved the chemical accuracy  for predicting a number of properties with their ML models by combining the atom and bond features of molecules  with global state features  before being updated during the iterative process\cite{dMPNN}.

Molecules are 3-dimensional multiconformational objects and hence it is natural to assume that they can be well represented by the nuclear coordinates %and charges 
as is the case of quantum mechanics based molecular simulations\cite{GOLLER20201702}. However, with coordinates the representation of molecules is non-invariant, non-invertible and non unique in  nature\cite{representations2} and hence not commonly used in conventional machine learning. In addition, the coordinates by itself do not carry information about the key attribute of molecules such as  bond types, symmetry, spin states, charge  etc in a molecule.  
Approaches/architectures have been proposed to create the robust, unique, invariant representations from nuclear coordinates using atom centered  Gaussian functions, tensor field networks, and more robustly by using representation learning techniques\cite{DTNN, SCHNET2, SchNet, SchNetPack, MEGNet, axelrod2020geom}.

 Chen et al.\cite{MEGNet} achieved the chemical accuracy  for predicting a number of properties with their ML models by combining the atom and bond features of molecules  with global state features of the molecules before being updated during the iterative process.  Robust representation of molecules can also be learned only from the nuclear charge and coordinates of molecules as demonstrated by Schutt et al\cite{SchNet, SchNetPack, DTNN}. 
Different variants (see Table 1) of message passing neural networks for representation learning have been proposed, with the main differences being how message are passed between the nodes and edges and how they are updated during the iterative process  using hidden states $h_v^{t}$. Hidden states at each nodes during message passing phase  are updated using

\begin{equation}
    m_v^{t+1} = \sum M_t(h_v^{t}, h_w^{t}, h_{vw}^{t}),~~~
h_v^{t+1} = S_t(h_v^{t}, m_v^{t+1})
\end{equation}

where $M_t$, $S_t$ are message and vertex update functions whereas $h_v^{t}$, $h_{vw}^{t}$ are node and edge features. The summation runs over all the neighbour of $v$ in the entire molecular graph. This information is used by a readout phase to generate the feature vector for molecule which is then used for property prediction.

These approaches however, require the relatively large amount of data and computationally intensive DFT optimized ground state coordinates for the desired accuracy, thus limiting their use for domains/data-sets lacking them. Moreover, representations learned from a particular 3D coordinate of a molecule fail to capture the conformer flexibility on its potential energy surface\cite{axelrod2020geom} thus requiring expensive multiple QM based calculations for each conformer of the molecule. Some work in this direction is based on semi-empirical DFT calculations to produce a database of conformers with 3D-geometry has been recently published\cite{axelrod2020geom}. This however does not provide any significant improvement in predictive power.
These methods in practice can be used with 
empirical coordinates generated from SMILES, using RDkit/chemaxon but still  require the corresponding ground state target properties for building a robust predictive modeling engine as well as to optimize the properties of new molecules with generative modelling.

%\textcolor{red}{One particular work in this direction was reported by d-MPNN et al., where they have used only SMILES string in QM9 dataset to generate atom, bond and global state attributes of the molecules. They however do not use the 3d-coordinates of the molecules. The main downside of their approach comes from the fact that, QM9 data have 133k molecules with nearly 50k unique SMILES strings. Their approach thus generates 50k unique input features to represent 133k molecules with 133k different target properties thus biasing the model, which is not ideal for robust ML model. }

Moreover, in these %MEGNet and SchNet
models, the cutoff distance is used to restrict the interaction among the atoms to the local environments only, hence generating local representations. In many molecular systems and for several applications, explicit non-local interactions is equally important\cite{short_range}. 
Long range interactions have been implemented in
convolutional neural networks,
however, are known to be inefficient in information propagation. Matlock et al.\cite{non-local_representations} proposed a novel architecture to encode non-local features of molecules in terms of efficient local features in
aromatic and conjugated systems using gated recurrent units. In their models, information is propagated back and forth in the molecules in the form of waves making it possible to pass the information locally while simultaneously travelling the entire molecule in a single pass. With the unprecedented success of learned molecular representations for predictive modelling, they are also adopted with success for generative models\cite{gschnet1, 3Dscaffolds}

\section{Predictive modeling}

Predictive modeling is the most widely studied area of applied machine learning in molecular modeling, drug discovery and medicine\cite{predictive3, predictive2, predictive1, predictive4, predictive5, predictive6, SchNet, DTNN, SchNetPack, dahal1}. % in the last decade and has been used to accurately predict the diverse properties in different science domains. 
Depending upon whether the ML architecture requires the pre-defined input representations as input features or can learn their own input representation by itself, %from the atomic number and corresponding nuclear coordinates, 
predictive modeling can be broadly classified into two sub-categories. The former is well covered in several recent review articles\cite{predictive3, predictive2, predictive1, predictive4, predictive5, predictive6}. We will focus only on the latter, which has been increasingly adopted in predictive machine learning recently with unprecedented accuracy for a range of properties and data-sets. A number of related approaches for predictive feature/property learning have been proposed in recent years under the umbrella term Graph Neural Network (GNN) \cite{duvenaud2015convolutional, FaberGraph, Fung2021} and extensively tested on different quantum chemistry benchmark datasets. GNN for predictive molecular modeling consists of two phases; representation learning and property prediction, integrated end-to-end in a way %not only 
to learn the meaningful representation of the molecules while simultaneously learning how to use the learned feature for the accurate prediction of properties. % Each of these two phase consist of sequence of layers. 
In the feature learning phase, atoms and bond connectivity information read from the nuclear coordinates or graph inputs are updated by passing through a sequence of layers for robust chemical encoding which are then used in subsequent property prediction blocks. % The learned feature can than be processed using dimensionality reduction techniques before using them in subsequent property prediction block.

In one of the first works on embedded feature learning, Schütt et al.\cite{DTNN} used the concept of many body Hamiltonians to devise the size of extensive, rotational, translational and permutationally invariant deep tensorial neural network architecture (DTNN) architecture for molecular feature learning and property prediction. Starting with the embedded atomic number and nuclear coordinates as input and after a series of refinement steps to encode the chemical environment, their approach learns the atom centered Gaussian-basis function as a feature which can be used to predict the atomic contribution for a given molecular property.
%They split the total property of molecule over atomic (single and pair-wise) contribution which when summed over gives total molecular property.
The total property of the molecule is the sum over the atomic contribution. They demonstrated chemical accuracy of 1 kcal mol$^{-1}$ in total energy prediction for relatively small molecules in QM7/QM9 dataset that contains only H, C, N, O, F atoms.%(single atom and pair-wise) %, predicted from their ML model.

\begin{table}[]
    \centering
%    \begin{adjustbox}{width=0.85\textwidth}
    \begin{adjustbox}{totalheight=\textheight-0\baselineskip, width=1\textwidth}

\begin{tabular}{p{0.1\textwidth}p{.5\textwidth}p{0.4\textwidth}p{0.4\textwidth}}

\hline\hline

Methods         &   Key feature       &     Advantage &       Drawbacks\\ \hline

\multirow{15}{*}{MPNN\cite{mpn-app}}  &
\begin{itemize}
  \item Message exchanged between the atoms  depends   only  on the feature of sending atom  and  corresponding   edge  features and is independent of  representation  of  atom  receiving the message
  \item Generate global representation of the molecule
  \item Predicted property of the molecule  is the function of global representations of the molecule
  \item Generate messages centered on the atoms
  
%  \item Message function
%  \begin{equation*}
%    m_v^{t+1} = \sum M_t(h_v^{t}, h_v^{t}, h_{vw}^{t}),
%h_v^{t+1} = S_t(h_v^{t}, m_v^{t+1})
%\end{equation*}
  \end{itemize}
  & 
  \begin{itemize}
  \item Achieved chemical accuracy in 11 out of 13   properties in QM9 data
  \item Performs well for intensive properties

\end{itemize}
  & 
  \begin{itemize}
  \item Requires optimized coordinates as a part of input features
  \item Including the state of message receiving atom (dubbed as pair message) increase the property prediction error
  \item The message passed from an atom A to atom B can be transmitted back to atom B resulting in noise
\end{itemize}
\\ \hline

%%% Second row

\multirow{8}{*}{d-MPNN\cite{dMPNN}} &
\begin{itemize}
  \item 	Learns molecular representation centered on bonds instead of atoms
  \item     Update on MPNN that combines learned representation with the prior known fixed atomic, bond and global molecular descriptors

  \end{itemize}
  & 
  \begin{itemize}
  \item Avoid noise resulting from message being passed along any path by using directed messages
  \item Use only SMILES string to generate input representation

\end{itemize}
%%%
  & 
  \begin{itemize}
  \item Requires optimized coordinates as a part of input features
\end{itemize}\\ \hline

%%% Third row

\multirow{10}{*}{SchNet\cite{SchNetPack}} &
\begin{itemize}
  \item 	Learns the atomistic representations of the molecules
  \item     Total property of the molecule is the sum  over the atomic contributions
  \item     Learns representations only by using the atomic number and geometry as atom and bond features respectively

  \end{itemize}
  & 
  \begin{itemize}
  \item Improve the performance on 8 out of 13 properties in QM9 data compared to MPNN
  \item Performs relatively well compared to MPNN for extensive properties
  \item Requires only nuclear charge and nuclear coordinates for learning input representations

\end{itemize}
%%%
  & 
  \begin{itemize}
  \item Relatively poor performance for intensive properties compared to MPNN
  \item Requires optimized 3D coordinates
\end{itemize}\\ \hline

%%% Fourth row

\multirow{15}{*}{MEGNet\cite{MEGNet}} &
\begin{itemize}
  \item 	Learns the atomistic representations of the molecules
  \item     Total property of the molecule is the sum  over the atomic contributions
  \item     Use several atomic and bond properties of the atom and bond as atom and bond features
  \item     Add the global state attribute of molecule in addition to atom and bond feature.

  \end{itemize}
  & 
  \begin{itemize}
  \item Improves the performance on all the extensive properties compared to MPNN and SchNet 
  \item Works equally well for molecules and solid

\end{itemize}
%%%
  & 
  \begin{itemize}
  \item Larger error for intensive properties compared to MPNN
  \item It calculate MAE errors for atomization energies of U0, U, H and G and compares with MAE on U0, U, H and G of SchNet
  \item Requires optimized 3D coordinates
\end{itemize}\\ \hline

%%% Fifth row

\multirow{4}{*}{SchNet-edge\cite{MPNN_edge}} &
\begin{itemize}
  \item 	Edge feature also depends upon the  features of the atom receiving the message

  \end{itemize}
  & 
  \begin{itemize}
  \item Improve the accuracy of the model over  SchNet/MPNN in all the properties in QM9 dataset 

\end{itemize}
%%%
  & 
  \begin{itemize}
  \item Requires optimized 3D coordinates

\end{itemize}\\ \hline\hline

\end{tabular}
\end{adjustbox}
            \caption{Benchmarks predictive modeling methods from literature with their key features, advantage and disadvantages}
\end{table}

Building on DTNN, Sch\"{u}tt et al.\cite{SchNetPack} also proposed a SchNet model, where the interactions between the atoms are encoded by using a continuous filter convolution layer before being processed by filter generating neural networks. They also expanded the predictive power of their model for  electronic, optical, and thermodynamic properties of molecules in the QM9 dataset compared to only total energy in DTNN achieving  state-of-the-art chemical accuracy in 8 out of 12 properties. They also improved on accuracy over a related approach of Gilmer et al.\cite{NMP}  known as message passing neural network (MPNN) on a number of properties except polarizability and electronic spatial extent. In contrast to the \textcolor{black}{SchNet/DTNN} model which learns atom-wise representation of molecule, MPNN learns the global representation of molecules
from the  atomic number, nuclear coordinates and other relevant bond-attributes and uses it for the molecular property prediction. MPNN is more accurate for the intensive properties ($\alpha$, %$<R^{2}>$.
$\braket{R^2}$) where the decomposition into individual atomic contribution is not required. % Their approach works equally well for predicting several properties of crystals.
Performance of SchNet is further improved by J{\o}rgensen et al.\cite{MPNN_edge} by making edge features inclusive of the atom receiving the message.

\begin{center}
\begin{table}[]
\begin{adjustbox}{width=\textwidth}
\begin{tabular}{lrrrrrr} 
\hline \hline
 Property                           & Units                     & MPNN          & SchNet-edge       &   SchNet          &   MegNet                       &  Target  \\ 
\hline 
 HOMO                               & eV                        & 0.043         &  \textbf{0.037}               & 0.041           &   0.038$\pm$0.001              & 0.043   \\ 
%\hline
 LUMO                               & eV                        & 0.037         &  \textbf{0.031}               & 0.034           &   \textbf{0.031$\pm$0.000}     & 0.043   \\ 
%\hline
 band gap                           & eV                        & 0.069         &  \textbf{0.058}               & 0.063           &   0.061$\pm$0.001              & 0.043   \\ 
%\hline
 ZPVE                               & meV                       & 1.500         &  1.490                        & 1.700           &   \textbf{1.400$\pm$0.060}     & 1.200   \\ 
%\hline
 dipole moment                      & Debye               & 0.030&  \textbf{0.029}               & 0.033           &   0.040$\pm$0.001              & 0.100   \\ 
%\hline
 polarizability                     & Bohr$^2$                  & 0.092         &  \textbf{0.077}               & 0.235           &   0.083$\pm$0.001     & 0.100   \\ 
%\hline
 R$^2$                              & Bohr$^2$                  & 0.180         &  \textbf{0.072}               & 0.073           &   0.265$\pm$0.001              & 1.200   \\
% \hline
 U$_0$                              & eV                        & 0.019         &  0.011$^\textbf{*}$           & 0.014           &   \textbf{0.009$\pm$0.000}$^\textbf{*}$     & 0.043   \\ 
%\hline
 U                                  & eV                        & 0.019         &  0.016$^\textbf{*}$           & 0.019           &   \textbf{0.010$\pm$0.000}$^\textbf{*}$     & 0.043   \\ 
%\hline
 H                                  & eV                        & 0.017         &  0.011$^\textbf{*}$           & 0.014           &   \textbf{0.010$\pm$0.000}$^\textbf{*}$     & 0.043   \\
%\hline
 G                                  & eV                        & 0.019         &  0.012$^\textbf{*}$           & 0.014           &   \textbf{0.010$\pm$0.000}$^\textbf{*}$     & 0.043   \\
%\hline
 C$_v$                              & cal (mol K)$^{-1}$        & 0.040         &  0.032                        & 0.033           &   \textbf{0.030$\pm$0.000}     & 0.050   \\
\hline
\hline

\end{tabular}
\end{adjustbox}
\caption{Mean absolute errors obtained from several benchmark methods on 12 different properties of molecules in QM9 dataset. Bold represents the lowest mean absolute errors among the models. $\textbf{*}$ represents  property are trained for respective atomization energies. Target corresponds to chemical accuracy for each properties desired from predictive models.}
\label{tab:my_label}
\end{table}
\end{center}

In another related model, Chen et al.\cite{MEGNet} proposed an integrated framework with unique feature update steps that work equally well for molecules and solids. They used several atom attributes \& bond-attributes and then combined it with the global state attribute to learn the feature representation of molecule. They claimed to out-perform SchNet model in 11 out of 13 properties including U0, U, H, and G in the benchmark QM9 dataset. However, they trained their model for respective atomization energies (P - n$_X$X$_p$, P = U0, U, H, and G) in-contrast to the parent U0, U, H, and G trained model of Schnet. 
A fair comparison of the model should be made between the similar quantities.
These models also demonstrated that a model trained for predicting a single property of molecules with GNN will always outperform the model optimized for predicting all the properties simultaneously. Other variants of MPNN are also published in the literature with slight improvements in accuracy for predicting some of the properties in the QM9 data set over the parent MPNN\cite{MPNN_edge, dMPNN}. The key features of a few benchmark models with their advantages and disadvantages are listed in Table 1. %One particular approach is of Jorgenson et al.\cite{MPNN_edge} where they extended SchNet model in a way that message exchanged between the atoms depends not only on the atom sending it but also on the atom receiving it.  
The comparison of mean absolute errors obtained from some of the benchmark models with their target chemical accuracy are reported in Table 2. 
This shows that appropriate ML models when used with proper representation of molecules, a well curated accurate data set, and a well sought state-of-the-art chemical accuracy from machine learning can be achieved.

%Graph bassed methods discussed  so far has been tested on small molecules consisting of lighter atoms (C, N, O, H, F) and mostly generate the local atomic representations. For larger molecules, methods have to include non-local interactions as well, vander-waals interactions. 

%They proposed a new architecture using  gated recurrent units to learn non-local features in
%aromatic and conjugated systems. 
%They demonstrated that non-local features can be represented with local variables by propagating information back and forth in the molecules in waves making it possible to pass the information locally while simultaneously travelling the entire molecule in a single pass. 

%Global state attribute of the molecules are combined with atom and bond features by megnet et al. and d-MPNN with improved performance on some of properties. 

%Map to series of structure

%\clearpage
%\textcolor{red}{talk about constraint: schnet conclusion} should capture the quantum mechanical interactions

%These models allows us to learn the complex molecular encoding only from the atomic coordinate and corresponding nuclear charge only.

\section{Inverse Molecular Design}
 
To achieve the long overdue goal of exploring a large chemical space, 
accelerated molecular design, and generation of molecules with desired  properties, inverse design is unavoidable. 
It is generally known that a molecule should have specific functionalities for it to be a effective therapeutic candidate against a particular disease, but in many cases new molecules that host such functionalities are not easily known with a direct approach. %As an example, it is well established in battery community that, an ideal cathode materials should provide large voltage, large energy storage capacity, long life and safe to operate, but such ''ideal" cathode material is not known yet limiting the use of batteries in large energy demanding industries such as in  electric vehicles, aeroplanes etc. 
Furthermore, the pool where such molecules may exist is astronomically large\cite{space1, space, COLEY2020} (approx. 10$^{60}$ molecules), making it impossible to explore each of them by  quantum mechanics based simulations or experiment.

In such scenarios, inverse design is of significant interest  %with the potential of revolutionizing the material discovery process. Inverse design
where the focus is on %the design and development of
quickly identifying novel molecules with desired properties, in contrast to the conventional, so called direct approach where known molecules are explored for different properties.  In inverse design, we start with the initial data set for which we know the structure and properties, map this to a probability distribution and then use it to generate new, previously unknown candidate molecules with desired properties very efficiently. Inverse design uses optimization and search algorithms\cite{zunger2018inverse, kuhn1996inverse} for the purpose %to learn functionality-surface mapping thus identifying the structures with desired functionality. Inverse design
and by itself can accelerate the lead molecule discovery process, %novel material prediction process 
which is the first step for any drug development. This paradigm holds even more promise when used in a closed loop with synthesis, characterization and different test tools in such a way that each of these steps receives and transmits feedback concurrently, thus improving each other over time.  This has shown some promise recently by substantially reducing the timeline for commercialization of molecules from its discovery to days, which is otherwise known to span over a decade in most cases. In one recent work, Zhavoronkov et al.\cite{DDR1_21_days} designed, developed and tested a workflow that integrates deep reinforcement learning with experimental synthesis, characterization and test tools for \textit{de novo} design of drug molecules as potential inhibitors of discoidin domain receptor-1 in 21 days.
Such a paradigm shift in the design of drugs is possible only because of recently developed deep generative model architectures.
Here, we briefly discuss some of the breakthrough architectures along with the recent applications in drug discovery.

%Generative models are used for the purpose which are trained to  reproduce realistic samples from the distribution of data. In generative molecular modelling, sequential SMILES string and graph based representations molecules are mostly used. The main challenge for generative modeling is optimization of latent space, which relied mainly on  reinforcement learning, continuous optimization in the latent space learned by variational autoencoders, or molecular translation.  

Variational autoencoders\cite{VAE} (VAE)
and its different variants have been extensively used for generating small  molecules  with optimal physico-chemical and biological properties.  
VAE consist of encoder and decoder network, where the encoder functions as a compression tool for compressing high-dimensional discrete molecular representations to a continuous vector in low-dimensional latent space, whereas the decoder recreates the  original molecules from the compressed space. Within VAE, recurrent neural networks (RNN)\cite{RNN} and convolution neural networks (CNN)\cite{CNN} are commonly used as encoding networks whereas several RNN based architectures such as GRU, LSTM are used as decoder network. RNN independently has also been used to generate molecules.
%used for dimensionality reduction, feature extraction, denoising applications, among other they are mostly popular for generative modelling.
Bombarelli et al.\cite{VAE} first used VAE's to generate molecules, in the form of SMILES strings, from latent space while simultaneously predicting their properties. For property prediction, they coupled the encoder, decoder network with the predictor network which uses vector from latent space as an input. SMILES strings generated from their VAE's do not always correspond to valid molecules. To improve on this, Kusner et al.\cite{GVAE} proposed a variant of VAE known as grammar VAE  that imposes constraint on SMILES generation by using  context free grammars rules. Both of these works employed string based molecular representations. % to SMILES string for encoding and decoding. % which constrain it to generate only syntactically-valid SMILES. %GVAE significantly increased the  generation of valid molecules. 
More recent works have focused on using molecular graphs as input and output for variational auto-encoders\cite{CGVAE} using
%Adding to Gomez works, graph based auto-encoders were used by ...
 different variants of VAE  among others\cite{CGVAE, GVAE, JT-VAE--3}, %conditional variational autoencoder,
 stacked auto-encoder, semi-supervised deep autoencoders, adversial autoencoder, Junction Tree Variational Auto-Encoder (JT-VAE) for generating molecules for drug discovery. %JT-VAE partition molecular graph into sub-graphs made up of %valid  components such as 
 %atoms, bonds, ring and generate a junction tree structured object  and encode molecule in two complementary latent space, one for molecular graph and another for corresponding junction tree.   
 In JT-VAE\cite{JT-VAE--3}, tree-like structures are generated from the valid sub-graph components of molecules and encoded along with a full graph to form two complementary latent spaces, one for molecular graph and another for the corresponding junction tree. These two spaces are then used for hierarchical decoding, generating  100 \% valid small molecules. Further improvement on this includes using JT-VAE in combination with auto-regressive and graph-to-graph translation methods for valid large molecule generation\cite{Graph-to-Graph_Translation}.

 %https://arxiv.org/pdf/1907.01632.pdf

Generative adversarial network (GAN) are another class of NN popular  for generating molecules\cite{dcGAN, Mol-GAN, Kadurin-GAN-2}. They consist of generative and discriminative models that work in coordination with each other where the generator is trained to generate a molecule and the discriminator is trained to check the accuracy of the generated molecules. Kadurin et al.\cite{Kadurin-GAN-2} successfully first used the GAN architecture for \textit{de novo} generation of molecules  with anti-cancer properties, where they demonstrated higher flexibility, %in producing molecular fingerprints, 
more efficient training, and processing of a larger data set compared to VAE. % using SMILES input representation. 
However, it uses unconventional binary chemical compound feature vectors and  %Compounds in PubChem were screened with the output fingerprints to identify candidates
requires cumbersome validation of output fingerprints against the PubChem
chemical library. Guimaraes et al.\cite{guimaraes-organ} and Sanchez-Lengeling et al.\cite{Sanchez-Lengeling2017-organic} used sequence based Generative Adversarial Network in combination with reinforcement learning for molecule generation, where they bias the generator to produce molecules with desired properties. The  works of Guimaraes et al. and Sanchez-Lengeling et al. suffer from several  issues associated with GAN, including mode collapse during training, among others. Some of these issues can be eliminated by %further extension to their work such as by 
using the reinforced adversarial neural computer method\cite{RANC}, which extends their work. Similar to VAE's, GAN's have also been used for molecular graph generation, which is considered more robust compared to SMILES string generation. Cao et al.\cite{Mol-GAN}  non-sequentially and efficiently generated the molecular graph of small molecules with high validity and novelty from jointly trained GAN and Reinforcement learning architectures. Maziarka et al.\cite{Graph-to-Graph_Translation} proposed a method for graph-to-graph translation, where they generated 100~\% valid molecules identical with the input molecules, but with different desired properties. Their approach relies on the latent space trained for JT-VAE and a degree of similarity of the generated molecules
 to the starting ones can be tuned. % via a hyperparameter. 
Mendez-Lucio et al.\cite{Mendez-Lucio2018} proposed conditional generative
adversarial networks to generate molecules that produce
a desired biological effect at a cellular level, thus bridging systems biology and molecular design. % without any previous target annotation of the training compounds as long as the gene expression signature of the desired state is provided.
deep convolution NN based GAN\cite{dcGAN} was used for \textit{de novo} drug design targeting types of cannabinoid receptors.

Generative models such as GAN's, RNN, VAE have been used together with reward-driven and dynamic decision making, reinforcement learning (RL) techniques in many cases with unprecedented success in generating molecules. %The unprecedented success come from their dynamic decision making strength which uses the estimate of action to improve the performance of the model.
Popova et al.\cite{Popova2018} recently used deep-RL for \textit{de novo} design of molecules with desired  hydrophobicity or %physico-chemical and biological properties.
 inhibitory activity against Janus protein kinase 2. They  trained generative and predictive model separately first and then  trained both together using a RL approach by biasing the model for generating molecules with desired properties.
In RL, an agent, which is a neural network takes actions to maximize the desired outcome by exploring the
chemical space and taking actions based on the reward, penalties, and policies setup to maximize the desired outcome. % \cite{https://advances.sciencemag.org/content/advances/4/7/eaap7885.full.pdf}
Olivecrona et al.\cite{olivecrona2017molecular} trained a policy-based RL model for generating the bioactives against dopamine receptor type 2, and generated molecules with more than 95 \% active molecules. Furthermore, taking an example of the drug Celecoxib, they demonstrated that,  RL can generate structure similar to Celecoxib  even when no Celecoxib was included in the training set. \textit{de novo}  drug design has so far focused only on generating structures that satisfy one of the several required criterion when used as drug.  Stahl et al. \cite{parameter-RL} proposed a  fragment-based RL approach employing
 an actor-critic model, for generating more than 90 \% valid molecules while optimizing multiple properties.  Genetic algorithms (GA) have also been used for generating molecules while optimizing their properties\cite{GA1, GA2, GA3, GA4}. GA based models suffer from stagnation while being trapped in at the regions of local optima\cite{trap_GA}. One notable work  alleviating these problems is by Nigam et al.\cite{selfiesApp1} where they hybridize GA and deep neural network to generate diverse molecules while outperforming related models in optimization.

All of the generative models discussed above generate molecules in the form of 2D graphs, or SMILES strings. Models to generate molecules directly in the form of 3D coordinates have also recently gained attention\cite{3D, 33D, gschnet}. Such generated 3D coordinates can be directly used for further simulation using quantum mechanics or by using docking methods. One of such first models is proposed by Niklas et al.\cite{gschnet} where they generate 3D coordinates of small molecules with light atoms (H, C, N, O, F). They then use the 3D coordinates of the molecules to learn representation to map it to a space which is then used to generate 3D coordinates of the novel molecules. Building on this for a drug discovery application, we recently proposed a model\cite{3Dscaffolds} to generate 3D coordinates of molecules while always preserving desired scaffolds. This approach has generated synthesizable drug-like molecules that show high docking score against the target protein. Other scaffolds based models to generates molecules in the form of 2D-graphs/SMILES strings are also published in the literature\cite{scaffold3, scaffold1, scaffold2, scaffold4, scaffold5}.

Recently, with the huge interest in the development of architecture and algorithms required for quantum computing, quantum version
of generative models such as the quantum auto-encoder\cite{Quantum-autoencoder} and quantum GANs\cite{Quantum-ML} have been proposed which carry huge potential, among others, for drug discovery. %, among other promises. %along with challenges in drug discovery. 
The preliminary proof of concept work of Romero et al.\cite{Quantum-ML, Quantum-autoencoder} shows that it is possible to encode and decode molecular information using a quantum encoder, demonstrating generative modeling is possible with quantum VAE's and more work especially in the development of supporting hardware architecture is required in this direction.

%One of the main challenge with generative model is the failure of generated molecules in experimental realization/synthesis. To account this, we proposed scaffold based 3D generative model which generates 3D coordinates of drug like molecules building from the desired scaffolds. Our model generates synthetically feasible drug like molecule

\section{Conclusions and Future Perspectives}

The success of current ML approaches depends upon how accurately we can represent a chemical structure for a given model. 
Finding a robust, transferable, interpretable, and easy to obtain representation which obeys the fundamental physics and chemistry of molecules that work for all different kind of applications  is  a critical task.  If available, this would save lot of resources, while increasing the accuracy and flexibility of molecular representations. Efficiently using such representations with robust and reproducible ML architectures will provide predictive modeling engine. Once a desired accuracy for diverse molecular systems for a given property prediction is achieved, it can routinely be used as an alternative to expensive QM based simulations or experiments.
In the chemical and biological sciences, a main bottleneck for deploying ML models is the lack of sufficiently curated data under similar conditions that is required for training the models. Finding architecture that works  consistently well enough for relatively small amount of data is equally important. Strategies such as active learning  (AL) and transfer learning (TL) are ideal for such scenarios 
 to tackle  problems\cite{AL-training_data, AL-drugs-ist, AL-HIV, BRADSHAW-AL,TL_images}. Graph based methods for end-to-end feature learning and predictive modeling so far have been successfully used on small molecules consisting of lighter atoms. For larger molecules, robust representation learning and molecule generation parts must include non-local interactions such as vander-waals and H-bonding  while building predictive and generative models. 

%Develop a scalable platform for molecular inverse design to radically accelerate the discovery of drugs, proteins and enzymes, industrial chemicals, biofuels, polymers, and bulk and composite materials.

% Initially focused on the design of antivirals against corona viruses, the infrastructure, algorithms, and software we build will be suitable for application to design challenges in other domains. Amongst other commercial applications, we aim to shorten end-to-end antiviral development and optimization times to less than four weeks upon the identification of a novel zoonotic transmission event -- prior to the next pandemic. More broadly, we aim to build the infrastructure required for proactive therapeutic development in advance of disease emergence -- powered by third-wave AI.

Equally important is to develop and tie a robust, transferable, and scalable state-of-the-art platform for inverse molecular design in a closed loop with predictive modeling engine to accelerate therapeutic design ultimately reducing the cost and time required.
Many of the inverse ML models used for inverse design use single bio-chemical activity as the criteria to measure the   success of generated  candidate therapeutic, which is in-contrast to real  clinical trial, where small molecule therapeutics are optimized for several bio-activities simultaneously. CAMD workflow should be designed in a way to optimize  multiple objective functions while generating and validating therapeutic molecules. Validation of newly generated lead molecules for a given drug usage, if done by experiments or quantum mechanical simulations, is an expensive task for all generated lead molecules. Ways to auto-validate molecules (using an inbuilt robust predictive model) would be ideal to save resources. In addition, CAMD workflows should be able to quantify uncertainty associated with it using statistical measures. For an ideal case, such uncertainty should decrease over the time as it learns from its own experience in series of closed loop.

CAMD workflows are generally built and trained with a specific goal in mind. Such workflows need to be re-configured and re-trained to work for different objective in therapeutic design and discovery.  \textcolor{black}{Design and build a single automated CAMD setup for multiple experiment (multi-parameter optimization) in a kind of transfer learning fashion is a challenge. }
It would be particularly helpful for the domains where a relatively small amount of data exist. Having such a CAMD infrastructure, algorithm and software stack would speedup end-to-end antiviral lead design and optimization for any future pandemics like Covid-19.

\section{Acknowledgement}
This research was supported by the DOE Office of Science through the National Virtual
Biotechnology Laboratory, a consortium of DOE national laboratories focused on response to COVID-19, with funding provided by the Coronavirus CARES Act. Additional support was provided by the Laboratory Directed Research and Development Program at Pacific Northwest National Laboratory (PNNL). PNNL is a multiprogram national laboratory operated by Battelle for the DOE under Contract DE-AC05-76RLO 1830. C. Computing resources was supported by the Intramural program at the William R. Wiley Environmental Molecular Sciences Laboratory (EMSL; grid.436923.9), a DOE Office of Science User Facility sponsored by the Office of Biological and Environmental Research and operated under Contract No. DE-AC05-76RL01830.

%\end{acknowledgement}

%\section*{References}

\bibliography{mybibfile}

\end{document}